\documentclass{article}
\usepackage{ijcai22}
\usepackage{times}
\usepackage{soul}
\usepackage{url}
\usepackage[hidelinks]{hyperref}
\usepackage[utf8]{inputenc}
\usepackage[small]{caption}
\usepackage{graphicx}
\usepackage{amsmath}
\usepackage{amsthm}
\usepackage{booktabs}
\usepackage{algorithm}
\usepackage{algorithmic}
\def\ie{{\em i.e.}}
\def\eg{{\em e.g.}}

\def\vs{\emph{vs.}}
\usepackage{diagbox}
\usepackage{amssymb}
\usepackage{multirow}

\title{MA-ViT: Modality-Agnostic Vision Transformers for Face Anti-Spoofing}

\author{
Ajian Liu$^{1,2}$\and
Yanyan Liang$^1$\thanks{Corresponding author}. 
$^1$School of Computer Science and Engineering, Faculty of Innovation Engineering, Macau University of Science and Technology, Macau \\
$^2$CBSR\&NLPR, Institute of Automation, Chinese Academy of Sciences, Beijing, China
\emails
ajianliu92@gmail.com,
yyliang@must.edu.mo
}

\begin{document}
\maketitle

\begin{abstract}
	The existing multi-modal face anti-spoofing (FAS) frameworks are designed based on two strategies: halfway and late fusion. However, the former requires test modalities consistent with the training input, which seriously limits its deployment scenarios. And the latter is built on multiple branches to process different modalities independently, which limits their use in applications with low memory or fast execution requirements. In this work, we present a single branch based Transformer framework, namely Modality-Agnostic Vision Transformer (MA-ViT), which aims to improve the performance of arbitrary modal attacks with the help of multi-modal data. Specifically, MA-ViT adopts the early fusion to aggregate all the available training modalities’ data and enables flexible testing of any given modal samples. Further, we develop the Modality-Agnostic Transformer Block (MATB) in MA-ViT, which consists of two stacked attentions named Modal-Disentangle Attention (MDA) and Cross-Modal Attention (CMA), to eliminate modality-related information for each modal sequences and supplement modality-agnostic liveness features from another modal sequences, respectively. Experiments demonstrate that the single model trained based on MA-ViT can not only flexibly evaluate different modal samples, but also outperforms existing single-modal frameworks by a large margin, and approaches the multi-modal frameworks introduced with smaller FLOPs and model parameters.
\end{abstract}

\section{Introduction}
Face Anti-Spoofing (FAS) aims to strengthen the face recognition system from a variety of presentation attacks (PAs)~\cite{Boulkenafet2017OULU,liu2018learning,liu20163d}. It has become an increasingly concern~\cite{zhang2020face,liu2020disentangling,liu2021face,liu2021contrastive,liu20213d} due to its wide applications in face payment, phone unlocking, and self-security inspection.
\begin{figure}[t]
	\centering
	\includegraphics[width=1.0\linewidth]{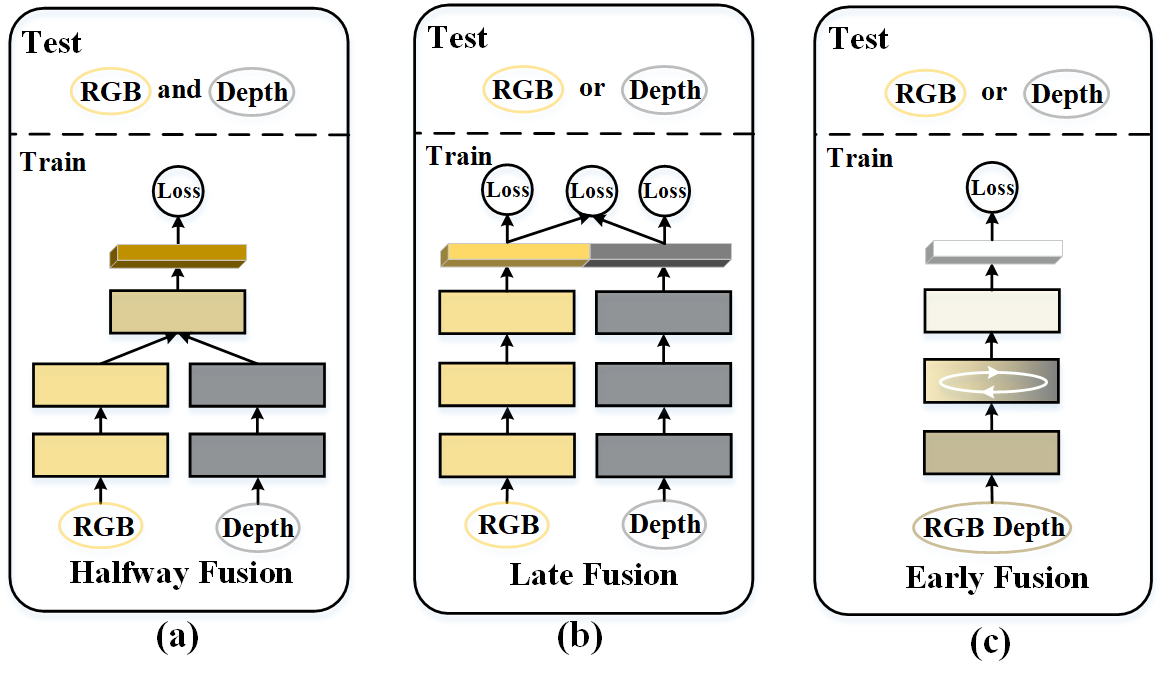}
	\caption{Comparison with existing multi-modal fusion strategies. (a) Halfway Fusion: it fuses multi-modal information by merging independent modal branches and requires the test modalities to be consistent with the training modalities. (b) Late Fusion: it fuses the multi-modal information at the decision level and requires complete modal branch for each input modality. (c) Early Fusion: it aggregates all available multi-modal data as input to realize flexible modal testing in a single branch structure. }
	\label{fig:fusion_strategy}
\end{figure}

With the increasingly advanced presentation attack instruments (PAIs) and acquisition sensors, face presentation attack detection (PAD) algorithms are also expanded from RGB spectrum~\cite{liu2018learning,shao2019multi,george2019deep,liu2019multi,yu2020searching} to multi-spectrum to explore more reliable spoofing traces, which can be divided into halfway fusion framework~\cite{zhang2020casia,parkin2019recognizing,george2019biometric} and late fusion framework~\cite{george2021cross}. As shown in Fig.~\ref{fig:fusion_strategy} (a), the halfway fusion is one of the commonly used fusion strategy, which combines the sub-networks of different modalities at a later stage via the feature concatenation. Although the halfway fusion can improve the robustness of PAD systems with the help of multi-modal information, their design concept is not intelligent enough, which requires the consistent training and testing modalities. If any modality disappears during testing, these methods would fail to distinguish live \vs fake faces and result in poor performance. In order to freely test any modal samples, as shown in Fig.~\ref{fig:fusion_strategy} (b), the late fusion strategy retains a specific branch for each modality to capture different modal information independently, and fuses the multi-modal information at the decision level (prediction embeddings or scores) to retain the function of any modal sample can be tested. However, these models may be slow to execute or large to store, limiting their use in applications or platforms with low memory or fast execution requirements, \eg, mobile phones. To sum up, it is a significant work to design a framework can combine relevant information from the different modalities without introducing additional parameters (or a small number of acceptable parameters) to improve the performance over using only one modality.

Whether it is halfway or late fusion, the common is that different modal inputs need to be mapped to a higher semantic layer in advance for knowledge fusion. Inspired by the success of Transformer architecture in natural language tasks, several recent works have also applied transformer-based fusion between image-text, image-audio, and image-text-audio with self-attention. In this work, as shown in Fig.~\ref{fig:fusion_strategy} (c), we adopt early fusion strategy by projecting all inputs from different modalities into a joint embedding space at input level and capture intra- and inter-modality interactions homogeneously within a pure transformer architecture. Compared with ConvNet-based framework, Transformer can not only capture local spoofing traces and establish their dependence on the patch tokens for a long time, but also can fuse multi-modal information directly without aligning different modal features in advance. Therefore, abandoning the CNNs to mine spoofing traces, we adopt a single branch transformer framework to aggregate all the available training modalities’ data and introduce the Modality-Agnostic Transformer Block (MATB) which consists of two cascaded attentions named Modal-Disentangle Attention (MDA) and Cross-Modal Attention (CMA) to improve the performance of any single-modal attacks at the inner- and inter modal levels, respectively. Specifically, the former eliminates modal-related information in the classification token under the guidance of a introduced modality token, and the latter supplements the modality-agnostic liveness information from other modalities. To sum up, the main contributions of this paper are summarized as follows:
\begin{itemize}
	\setlength{\itemsep}{1.0pt}
	\item
	We present a novel Modality-Agnostic Vision Transformer (MA-ViT) framework to improve the performance of any single modal FAS system with the help of available multi-modal training data. Compared with the previous multi-modal fusion frameworks, our approach has the advantages of flexible testing of different modal samples and less introduction of additional model parameters.
	\item
	We develop the Modality-Agnostic Transformer Block (MATB) in MA-ViT with two effective attentions, namely Modal-Disentangle Attention (MDA) and Cross-Modal Attention (CMA), to achieve the modality-agnostic and complementary liveness features learning, respectively.
	\item
	Extensive experiments are conducted and demonstrate that the proposed MA-ViT can improve the performance of a single-modal system with the help of multi-modal data, with only acceptable FLOPs and model parameters are increased.
\end{itemize}

\section{Modality-Agnostic ViT (MA-ViT)}
\subsection{Overall Framework}
\begin{figure*}[t]
	\centering
	\includegraphics[width=1.0\linewidth]{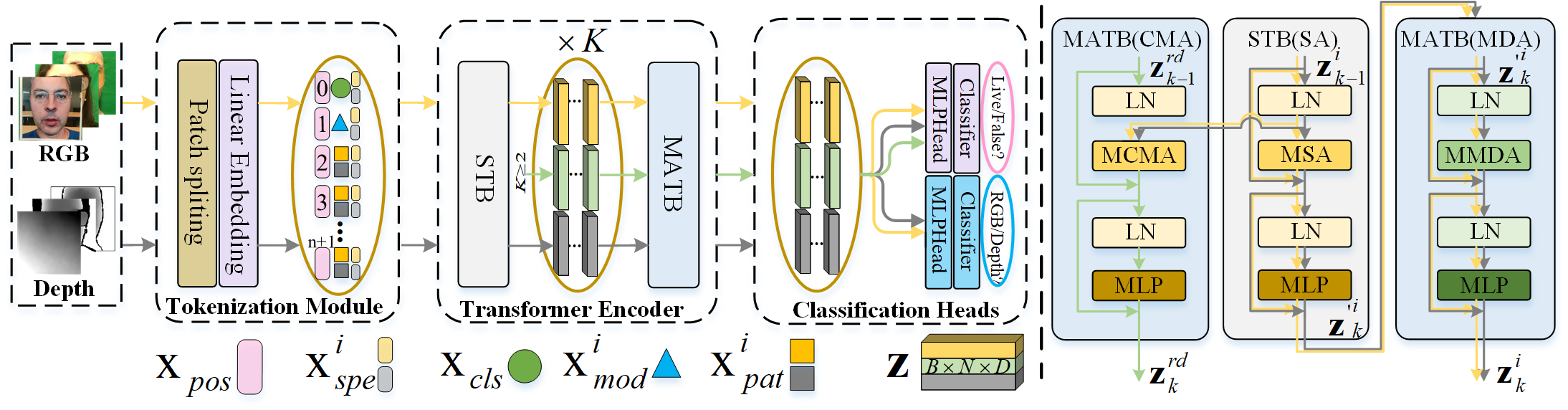}
	\caption{The architecture of Modality-Agnostic Vision Transformer (MA-ViT). It is built on a single ViT branch and consists of Multi-Modal Tokenization Module (Eq.~\ref{Eq:z0}), Transformer Encoder, and Modal Shared Classification Heads. (Eq.~\ref{Eq:zL}). A completed transformer encoder contains $K$ transformer block, which consists of Standard Transformer Block (STB) and Modality-Agnostic Transformer Block (MATB). Specifically, the STB processes all modal sequences (yellow and black gray arrows) independently and outputs them into MATB (MDA). And the MATB (CMA) takes all modal sequences as inputs and outputs the cross-modal sequences (green arrow).}
	\label{fig:architecture}
\end{figure*}

As shown in Fig.~\ref{fig:architecture}, our MA-ViT is built on the standard ViT~\cite{dosovitskiy2020image} which consists of tokenization module, transformer encoder, and classification head. In addition, a learnable modality token is introduced to summarizes the modal information from patch tokens. A proposed MATB is inserted after Standard Transformer Block (STB) to disentangle the modal information and supplement the liveness features for the classification token, respectively. Let $\mathbf{I}^{i}$ represents the input in a training batch, where $i$ can be $r$ and $d$ for the RGB and Depth samples, and also can be easily extended to other modalities.

{\textbf{Multi-Modal Tokenization Module.}}
MA-ViT first splits any modal input $\mathbf{I}$ into a sequence of non-overlapping patches by a patch splitting module and then linearly projecting patches into tokens $\mathbf{x}_{pat} \in \mathbb{R}^{n\times D}$, where $n$ is the number of resulting patches, and $D$ is vector size through all of MA-ViT layers. Similar to BERT and DeiT~\cite{touvron2021training}, a learnable class token ($\texttt{CLS}$) $\mathbf{x}_{cls}=\mathbf{z}_{0,cls} \in \mathbb{R}^{1\times D}$ and a modality token ($\texttt{MOD}$) $\mathbf{x}_{mod}=\mathbf{z}_{0,mod} \in \mathbb{R}^{1\times D}$ are concatenated to the patch tokens, who serves as the image and modality agents for classification. Due to different modal images are collected by fixed position multi-spectral sensor, they share location information and exclusive spectrum space. Therefore, position embeddings $\mathbf{x}_{pos} \in \mathbb{R}^{(n+2)\times D}$ and spectrum embeddings $\mathbf{x}_{spe} \in \mathbb{R}^{(n+2)\times D}$ are added to each token to retain positional and modal informations. The tokenization process is expressed as follows:
\begin{equation}\label{Eq:z0}
	\mathbf{z}^{i}_{0}=[\mathbf{x}_{cls}||\mathbf{x}^{i}_{mod}||\mathbf{x}^{i}_{pat}]+(\mathbf{x}_{pos} +\mathbf{x}^{i}_{spe}), \mathbf{z}^{i}_{0}\in \mathbb{R}^{N\times D}
\end{equation}
where $||$ means token concatenation. $i\in \left \{ r,d \right \}$ and $N=n+2$ represent the modal type and the total number of tokens. respectively. The resulting sequence $\mathbf{z}_{0}$ serves as input to the following transformer encoder.

{ \textbf{Transformer Encoder with MATBs.}}
After tokenization module, the sequences $\mathbf{z}^{i}_{0}$ ($i\in \left \{ r,d \right \}$) from all modalities are mixed through STB and MATB, which constitute one transformer block of our MA-ViT. The complete transformer encoder consists of $K$ transformer blocks by repeatedly stacking. As illustrated in Fig.~\ref{fig:architecture}, a MATB with MDA receives the output of the previous STB as input, and with CMA receives the current intra-modal sequences ($\mathbf{z}^{i}$) and the previous cross-modal sequence ($\mathbf{z}^{rd}$) as input. Similar to Multi-headed Self-Attention (MSA) in STB, a Layer normalization (LN) is applied before each module, and a residual shortcut is applied after each module. Specifically, the processing of $k$-th (where $k=1,...,K$) Transformer block can be expressed as (omit MLP operation):
\begin{equation}\label{Eq:matb}
	\begin{split}		
		\mathbf{z'}^{i}_{k}&=\mathrm{MSA}(\mathrm{LN}(\mathbf{z}^{i}_{k-1}))+\mathbf{z}^{i}_{k-1} \\
		\mathbf{z}^{rd}_{k}&=\mathrm{MCMA}(\mathrm{LN}(\mathbf{z}^{r}_{k-1}), \mathrm{LN}(\mathbf{z}^{d}_{k-1}))+\mathbf{z}^{rd}_{k-1} \\
		\mathbf{z}^{i}_{k}&=\mathrm{MMDA}(\mathrm{LN}(\mathbf{z'}^{i}_{k}))+\mathbf{z'}^{i}_{k} \\
	\end{split}
\end{equation}
where MSA and MCMA (Multi-headed CMA) and the following MLP weights are shared. $\mathbf{z}^{i}_{k}$ and $\mathbf{z}^{rd}_{k}$ (cross-modal sequence) are the output of MATB (MDA) and MATB (CMA). 

{ \textbf{Modal Shared Classification Heads.}}
See from Fig.~\ref{fig:architecture}, we provide two modal shared classification heads for all sequences to meet the requirements of any modal sample can be tested, which are supervised and jointly optimized by binary cross-entropy (BCE). The output sequences at the $K$-th block in Transformer encoder $\mathbf{z}_{K,cls}$ and $\mathbf{z}_{K,mod}$ are served as the agents for liveness and modality classification, respectively. The total loss $L^{total}$ to minimize is given as:
\begin{equation}\label{Eq:zL}
	\begin{split}		
		L^{i}_{cls}&=\mathrm{BCE}(\mathrm{MLP}^{cls}(\mathrm{LN}(\mathbf{z}_{K,cls}^{i})), y_{cls}), \\
		L^{i}_{mod}&=\mathrm{BCE}(\mathrm{MLP}^{mod}(\mathrm{LN}(\mathbf{z}_{K,mod}^{i})), y_{mod}), \\
		L^{i}_{total} &=L^{i}_{cls}+L^{rd}_{cls}+L^{i}_{mod}, i\in \left \{ r,d \right \}
	\end{split}
\end{equation}
where $L^{rd}_{cls}$ is the loss for cross-modal sequence $\mathbf{z}^{rd}_{K}$. The classification head is implemented by MLP with a single linear layer. $y$ is the ground truth for sample $\mathbf{I}$. In detail, $y_{cls}=0/1$ for attack/bonafide and $y_{mod}=0/1$ for RGB/Depth, respectively. 

\subsection{Modality-Agnostic Transformer Block.}  

{ \textbf{Multi-headed Modal-Disentangle Attention (MMDA).}} 
How to determine the modality-irrelated patch tokens is the primary task in MMDA. By analyzing the attention matrix, which is essentially a relevance maps, whose each row corresponds to a link for each token given the other tokens. In which the relevance map that corresponds to the $\texttt{CLS}$/$\texttt{MOD}$ token links each of the tokens to the $\texttt{CLS}$/$\texttt{MOD}$ token, and the strength of this link can be intuitively considered as an indicator of the contribution of each token to the classification.

\begin{figure}[t]
	\centering
	\includegraphics[width=0.9\linewidth]{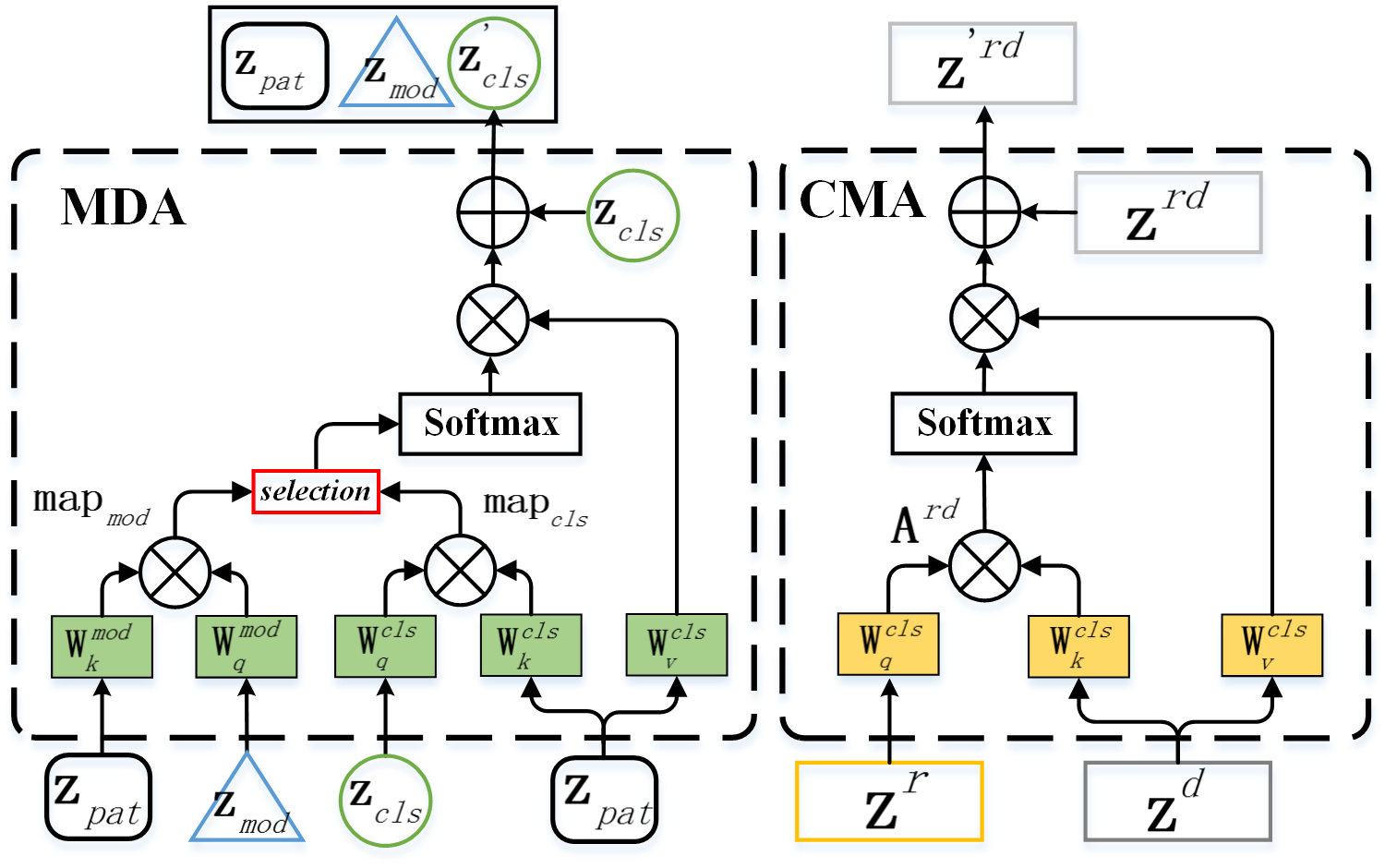}
	\caption{Implementation details of Modal-Disentangle Attention (MDA) and and Cross-Modal Attention (CMA).}
	\label{fig:attentions}
\end{figure}

See from the MATB (MDA) in Fig.~\ref{fig:architecture}, for any modal sequence $\mathbf{z}^{i}$, as shown in Eq.~\ref{Eq:mda} and Fig.~\ref{fig:attentions}, we first compute $\mathbf{q}_{mod}$ of $\texttt{MOD}$ token and $\mathbf{k}^{mod}_{pat}$ of patch tokens through two learnable parameters $\mathbf{W}^{mod}_{q}$ and $\mathbf{W}^{mod}_{k}$, respectively. Then, we compute the dot products of the $\mathbf{q}_{mod}$ with $\mathbf{k}^{mod}_{pat}$, divide each by $\sqrt{D/h}$, and apply a softmax function to obtain the modal relevance map $\mathbf{map}_{mod}$. Then, we identify the modality-irrelated (or weakly related) patch tokens by excluding strong links in $\mathbf{map}_{mod}$, which is completed by a threshold function $\Gamma_{\lambda }(\cdot )$. Specifically, $\Gamma_{\lambda}(\cdot )$ aims to find the modality-related patch tokens by thresholding the modal relevance map to keep $\lambda (\in  [0,1])$ proportional mass. It outputs a mask matrix $\textbf{M}$ with values of 1 and 0, where 1/0 means to modality-related/-irrelated the patch tokens of the corresponding position. 

In order to disentangle the modal information for classification, we take the mask matrix $\textbf{M}$ as indicators to discard the modality-related patch tokens. As shown in Eq.~\ref{Eq:mda}, we first compute $\mathbf{q}_{cls}$ of $\texttt{CLS}$ token, $\mathbf{k}^{cls}_{pat}$ and $\mathbf{v}^{cls}_{pat}$ of patch tokens through three learnable parameters $\mathbf{W}^{cls}_{q}$, $\mathbf{W}^{cls}_{k}$ and $\mathbf{W}^{cls}_{v}$, respectively. Then, we obtain the attention map $\mathbf{map}_{cls}$ in a similar way by $\mathbf{q}_{cls}$ and $\mathbf{k}^{cls}_{pat}$, while disconnecting the links with the modality-related patch tokens under the guidance of $\textbf{M}$. See from the Eq.~\ref{Eq:mda}, this process can be completed with a selection function $\Gamma'_{\textbf{M}}(\cdot)$ based on the index position of a given matrix $\textbf{M}^{i}$. Finally, we successively obtain the weights by a softmax function to $\mathbf{map}_{cls}$ and the output of MDA by the weighted sum over all values of patch tokens $\mathbf{v}^{cls}_{pat}$. This process is denoted as:
\begin{equation}\label{Eq:mda}
	\begin{split}		
		& [\mathbf{q}_{cls},\mathbf{k}^{cls}_{pat},\mathbf{v}^{cls}_{pat}]=[\mathbf{z}_{cls}\mathbf{W}^{cls}_{q}, \mathbf{z}_{pat}\mathbf{W}^{cls}_{k}, \mathbf{z}_{pat}\mathbf{W}^{cls}_{v}], \\
		& [\mathbf{q}_{mod},\mathbf{k}^{mod}_{pat}]=[\mathbf{z}_{mod}\mathbf{W}^{mod}_{q}, \mathbf{z}_{pat}\mathbf{W}^{mod}_{k}], \\
		& \mathbf{map}_{mod}=\mathbf{q}_{mod}(\mathbf{k}_{pat}^{mod})^{T}/\sqrt{D/h}, \\ 
		& \mathbf{map}_{cls}=\mathbf{q}_{cls}(\mathbf{k}_{pat}^{cls})^{T}/\sqrt{D/h},\\
		& \textbf{M}=\Gamma_{\lambda }(\mathbf{map}_{mod}), \\
		& \mathrm{MDA}(\mathbf{z}^{i})=\mathrm{softmax}[\Gamma'_{\textbf{M}}(\mathbf{map}_{cls})]\cdot \mathbf{v}^{cls}_{pat}
	\end{split}
\end{equation}
where $D$ and $h$ are the embedding dimension and number of heads, respectively. $\mathbf{W}_{q}$, $\mathbf{W}_{k}$ and $\mathbf{W}_{v}$ $\in \mathbb{R}^{D\times (D/h)}$. $\Gamma'_{\textbf{M}}(\cdot)$ is a selection function defined as:
\begin{equation}\label{Eq:mask}
	\begin{split}
		\Gamma'_{\textbf{M}}(\mathbf{A})=\left\{\begin{matrix}
			-\infty, \textbf{M}_{a,b}>0 & \\ 
			\mathbf{A}_{a,b}, \textbf{M}_{a,b}=0 & 
		\end{matrix}\right.	
	\end{split}
\end{equation}
Similar to~\cite{wang2021kvt}, we discard the modality-related tokens by setting the attention values in $\mathbf{map}_{cls}$ to small enough constant. MMDA is an extension of MDA in which we run $h$ mutual-attention operations in parallel. As shown in Fig.~\ref{fig:architecture} and Eq.~\ref{Eq:matb}, we apply a residual shortcuts after the MDA and concatenate with patch tokens $\mathbf{z}_{pat}$ to obtain a new sequences $\mathbf{z'}$:
\begin{equation}\label{Eq:mda_z'}
	\begin{split}		
		\mathbf{z'}^{i}_{cls}=\mathrm{MMDA}(\mathbf{z}^{i})+\mathbf{z}^{i}_{cls}, \mathbf{z'}^{i} &= [\mathbf{z'}^{i}_{cls}||\mathbf{z}^{i}_{mod}||\mathbf{z}^{i}_{pat}]
	\end{split}
\end{equation}
where $||$ means token concatenation, the output sequences $\mathbf{z'}^{i}$ will be used as the input sequences of the next stage.
    
{ \textbf{Multi-headed Cross-Modal Attention (MCMA).}}
How to effectively produce a stronger modality-agnostic representation with the help of multi-modal data is the primary task in MCMA. Inspired by CDTrans~\cite{xu2021cdtrans}, we summarize the common information of different modal sequences to construct cross-modal sequence output by MCMA, which is supervised by the same label. In other words, the MCMA further guides MSA to pay attention to modality-agnostic liveness features by sharing with its parameters.

See from the MATB (CMA) module in Fig.~\ref{fig:architecture}, it takes all modal sequences $\mathbf{z}^{r}$ and $\mathbf{z}^{d}$ as inputs and outputs cross-modal sequence $\mathbf{z}^{rd}$. Similar to MSA~\cite{dosovitskiy2020image}, as shown in Fig.~\ref{fig:attentions}, we first compute queries $\mathbf{q}^{r}$ of modal sequence $\mathbf{z}^{r}$, and keys $\mathbf{k}^{d}$, values $\mathbf{v}^{d}$ from another modal sequence $\mathbf{z}^{d}$, respectively. Then, the attention function $\textbf{A}^{rd}$ is computed on the set of queries $\mathbf{q}^{r}$ simultaneously with all keys $\mathbf{k}^{d}$. Finally, the outputs of CMA is a weighted sum over all values $\mathbf{v}^{d}$, denoted as:
\begin{equation}\label{Eq:cma}
	\begin{split}		
		& [\mathbf{q}^{r},\mathbf{k}^{d},\mathbf{v}^{d}]=[\mathbf{z}^{r}\mathbf{W}^{cls}_{q}, \mathbf{z}^{d}\mathbf{W}^{cls}_{k}, \mathbf{z}^{d}\mathbf{W}^{cls}_{v}], \\
		& \textbf{A}^{rd}=\mathrm{softmax}(\mathbf{q}^{r}(\mathbf{k}^{d})^{T}/\sqrt{D/h}), \\
		& \mathrm{CMA}(\mathbf{z}^{i})=[\textbf{A}^{rd} \cdot \textbf{v}^{d} || \textbf{A}^{dr} \cdot \textbf{v}^{r}], i\in \left \{ r,d \right \}
	\end{split}
\end{equation}
where the calculation process of $\textbf{A}^{dr}$ and $\textbf{v}^{r}$ is similar to that of $\textbf{A}^{rd}$ and $\textbf{v}^{d}$ by simply swapping the index $r$ and $d$. $\mathbf{W}_{q}$, $\mathbf{W}_{k}$ and $\mathbf{W}_{v}$ are shared with MSA. $||$ means concatenating cross-modal sequences along the batch dimension. Similar to MDA in Eq.~\ref{Eq:mda_z'}, we also apply a residual shortcuts after the CMA, and obtain the new cross-modal sequence $\mathbf{z'}^{rd}=\mathrm{CMA}(\mathbf{z}^{i})+\mathbf{z}^{rd}$ (The operation is not performed in the 1-th layer).

\section{Experiments} 
\subsection{Experimental Setup} 
{\flushleft \textbf{Datasets \& Protocols.}}
We use three commonly used multi-modal and a single-modal FAS datasets for experiments, including CASIA-SURF (MmFA)~\cite{zhang2019dataset}, CASIA-SURF CeFA (CeFA)~\cite{liu2021casia}, and WMCA~\cite{george2019biometric}. MmFA consists of $1,000$ subjects with $21,000$ videos and each sample has $3$ modalities, and provides a intra-testing protocol. CeFA covers $3$ ethnicities, $3$ modalities, $1,607$ subjects, and provides five protocols to measure the affect under varied conditions. We select the Protocol 1, 2, and 4 for experiments. WMCA contains a wide variety of 2D and 3D presentation attacks, which introduces 2 protocols: grandtest protocol which emulates the ``seen'' attack scenario and the ``unseen'' attack protocol that evaluates the generalization on an unseen attack. 

{ \textbf{Test Scenario Settings.}}
We consider two test scenarios. The first is the commonly used setting that the test modalities need to be consistent with the training stage. In this scenario, we evaluate the intra-testing performances of each dataset based on the provided protocols, and evaluate the robustness of our approach through cross-testing experiments between these datasets. The second is the flexible modal test scenario, which means the tester can provide any single-modal or any multi-modal paired samples.

{ \textbf{Evaluation Metrics.}}
In intra-testing experiments, Attack Presentation Classification Error Rate (APCER), Bonafide Presentation Classification Error Rate (BPCER), and ACER are used for the metrics. The ACER on testing set is determined by the Equal Error Rate (EER) threshold on dev sets for MmFA, CeFA, OULU, and the BPCER=$1\%$ threshold for WMCA. TPR(@FPR=$10^{-4}$) is also provided for MmFA. For cross-testing experiments, Half Total Error Rate (HTER) is adopted as the metric, which computes the average of False Rejection Rate (FRR) and the False Acceptance Rate (FAR), and the threshold computed in dev set using EER criteria.

{ \textbf{Implementation Details.}} 
Our models can be freely built on any version of ViT~\cite{dosovitskiy2020image}, which means that our MATB can be compatible with other improvements to ViT, such as DeiT~\cite{touvron2021training} and Swin~\cite{liu2021swin}. In our experiments, we adopt ViT-S/16 as the backbone through comparative experiments, which means the ``Small'' variant with $K=12$. 

We train our model with the ViT-S/16 backbone, which is initialized with weights provided by~\cite{touvron2021training}, and other newly added layers are randomly initialized. We resize all modal images to $224\times224$ and train all models with $50$ epochs via Adam solver. All models are trained with a batch seize of 8 and an initial learning rate of 0.0001 for all epochs. We set $\lambda=0.8$ in MDA according to comparative experiments. In the testing stage, for single-mdoal samples, we only activate the input interface of the corresponding modality (mainly the corresponding position and spectrum embeddings). As can be seen more clearly from Fig~\ref{fig:architecture}, the green line only outputs the results of multi-modal samples, while the yellow and gray lines output the results of corresponding single-modal samples respectively. 

\begin{table*}[]
	\centering
	\scalebox{0.90}{
		\begin{tabular}{|c|c|cccccccc|}
			\hline
			\multirow{2}{*}{Method} & \multirow{2}{*}{seen} & \multicolumn{8}{c|}{unseen}                                                                                                                                                                                                                                                             \\ \cline{3-10} 
			&                       & \multicolumn{1}{c|}{Flexiblemask}  & \multicolumn{1}{c|}{Replay}        & \multicolumn{1}{c|}{Fakehead}      & \multicolumn{1}{c|}{Prints}        & \multicolumn{1}{c|}{Glasses}        & \multicolumn{1}{c|}{Papermask}     & \multicolumn{1}{c|}{Rigidmask}     & mean$\pm$std        \\ \hline \hline
			MC-PixBiS               & 1.80                  & \multicolumn{1}{c|}{49.70}         & \multicolumn{1}{c|}{3.70}          & \multicolumn{1}{c|}{0.70}          & \multicolumn{1}{c|}{0.10}          & \multicolumn{1}{c|}{16.00} & \multicolumn{1}{c|}{0.20}          & \multicolumn{1}{c|}{3.40}          & 10.50$\pm$16.70     \\ \hline
			MCCNN-OCCL-GMM          & 3.30                  & \multicolumn{1}{c|}{22.80}         & \multicolumn{1}{c|}{31.40}         & \multicolumn{1}{c|}{1.90}          & \multicolumn{1}{c|}{30.00}         & \multicolumn{1}{c|}{50.00}          & \multicolumn{1}{c|}{4.80}          & \multicolumn{1}{c|}{18.30}         & 22.74$\pm$15.30     \\ \hline
			CMFL                    & 1.70                  & \multicolumn{1}{c|}{12.40}         & \multicolumn{1}{c|}{1.00}          & \multicolumn{1}{c|}{2.50}          & \multicolumn{1}{c|}{0.70}          & \multicolumn{1}{c|}{33.50}          & \multicolumn{1}{c|}{1.80}          & \multicolumn{1}{c|}{1.70}          & 7.60$\pm$11.20      \\ \hline
			ViT                     & 3.79                  & \multicolumn{1}{c|}{14.56}         & \multicolumn{1}{c|}{1.58}          & \multicolumn{1}{c|}{4.14}          & \multicolumn{1}{c|}{0.83}          & \multicolumn{1}{c|}{36.00}          & \multicolumn{1}{c|}{2.82}          & \multicolumn{1}{c|}{2.19}          & 7.93$\pm$13.80       \\ \hline
			MA-ViT                  & \textbf{1.45}         & \multicolumn{1}{c|}{\textbf{9.76}} & \multicolumn{1}{c|}{\textbf{0.93}} & \multicolumn{1}{c|}{\textbf{0.55}} & \multicolumn{1}{c|}{\textbf{0.00}} & \multicolumn{1}{c|}{\textbf{14.00}}          & \multicolumn{1}{c|}{\textbf{0.00}} & \multicolumn{1}{c|}{\textbf{1.46}} & \textbf{3.81$\pm$5.67} \\ \hline
		\end{tabular}
	}
	\caption{Comparison of ACER (\%) values on Protocol ``seen" and ``unseen" for the WMCA. Best results are bolded.}
	\label{tab:WMCA_results}
\end{table*}
\subsection{Fixed Modal Scenario Evaluations}
On the three multi-modal datasets, we compare with the baseline method ViT that removed the MATB from MA-ViT, and the previous state-of-the-art (SOTA) methods. 

{ \textbf{Intra-Testing Results.}} 
For MmFA, we compare with the benchmark SEF~\cite{zhang2019dataset} and the multi-scale version MS-SEF~\cite{zhang2020casia}, respectively. From the Tab.~\ref{tab:casiasurf_result}, we can observe that the performance of ViT is worse than SEF due to the lack of modal fusion ability. However, when equipped with MATB, MA-ViT improves the TPR from $44.33\%$ to $82.83\%$, reduces ACER from $6.51\%$ to $0.80\%$, and outperforms MS-SEF by a large margin, which uses ImageNet pretrain. 
\begin{table}[]
	\centering
	\resizebox{1.0\linewidth}{!}{
		\begin{tabular}{|c|c|c|c|c|}
			\hline
			Method     & APCER         & BPCER               & ACER             & TPR(@FPR=$10^{-4}$) \\ \hline \hline
			SEF           & 3.80             & 1.00                   & 2.40              & 56.80               \\ \hline 
			MS-SEF    & 2.30             & \textbf{0.30}     & 1.30               & 81.40                \\ \hline
			ViT           & 3.52              & 9.50                   & 6.51                 &   44.33                \\ \hline
			MA-ViT    & \textbf{0.78} & 0.83                   & \textbf{0.80}   &   \textbf{82.83}   \\ \hline
		\end{tabular}
	}
\caption{The results on MmFA. A large TPR(\%) and a lower ACER (\%) indicate better performance. Best results are bolded.}
\label{tab:casiasurf_result}
\end{table}

For CeFA, we compare our approach with the benchmark PSMM~\cite{liu2021casia} and the top three methods introduced in the challenge~\cite{liu2021cross} with the Protocol 4, \ie, BOBO, Super and Hulking. It can be seen from Tab.~\ref{tab:cefa_result} that the performance of ViT is superior to PSMM on Protocol 1 and Protocol 2 due to the pre-training model. Our MA-ViT further reduces the ACER on Protocol 1 and Protocol 2 to $1.10\%$ and $0.10\%$, and outperforms the Hulking and Super methods with ACER gains of $-0.04\%$ ($1.68\%$ vs. $1.64\%$) and $-0.57\%$ ($2.21\%$ vs. $1.64\%$), respectively.
\begin{table}[]
	\centering
	\resizebox{1.0\linewidth}{!}{
\begin{tabular}{|c|c|c|c|c|}
	\hline
	Pro.               & Method  & APCER(\%)              & BPCER(\%)              & ACER(\%)               \\ \hline \hline
	\multirow{3}{*}{1} & PSMM    & 2.40$\pm$0.60          & 4.60$\pm$2.30          & 3.50$\pm$1.30          \\ \cline{2-5} 
	& ViT     & 4.55$\pm$7.29          & 1.83$\pm$1.59          & 3.19$\pm$4.07          \\ \cline{2-5} 
	& MA-ViT  & \textbf{1.45$\pm$1.75} & \textbf{0.75$\pm$0.43} & \textbf{1.10$\pm$1.09} \\ \hline
	\multirow{3}{*}{2} & PSMM    & 7.70$\pm$9.00          & 3.10$\pm$1.60          & 5.40$\pm$5.30          \\ \cline{2-5} 
	& ViT     & 3.48$\pm$1.81          & 0.92$\pm$0.12          & 2.20$\pm$0.85          \\ \cline{2-5} 
	& MA-ViT  & \textbf{0.12$\pm$0.08} & \textbf{0.09$\pm$0.12} & \textbf{0.10$\pm$0.01} \\ \hline
	\multirow{6}{*}{4} & PSMM    & 7.80$\pm$2.90          & 5.50$\pm$3.00          & 6.70$\pm$2.20          \\ \cline{2-5} 
	& Hulking & 3.25$\pm$1.98          & 1.16$\pm$1.12          & 2.21$\pm$1.26          \\ \cline{2-5} 
	& Super   & \textbf{0.62$\pm$0.43} & 2.75$\pm$1.50          & 1.68$\pm$0.54          \\ \cline{2-5} 
	& BOBO    & 1.05$\pm$0.62          & \textbf{1.00$\pm$0.66} & \textbf{1.02$\pm$0.59} \\ \cline{2-5} 
	& ViT     & 20.44$\pm$17.12        & 7.42$\pm$3.96          & 13.92$\pm$10.45        \\ \cline{2-5} 
	& MA-ViT  & 2.10$\pm$1.47          & 1.17$\pm$0.38          & 1.64$\pm$0.89          \\ \hline
\end{tabular}
	}
	\caption{Evaluation results (\%) on the Protocol 1, 2, and 4 of CeFA dataset. Note that a lower ACER value indicates better performance. } 
	\label{tab:cefa_result}
\end{table}

In order to perform a fair comparison with prior methods, only RGB and Depth data in WMCA~\cite{george2019biometric} are used for intra-testing experiments. Tab.~\ref{tab:WMCA_results} presents the comparisons of ACER to the SOTA ConvNet-based methods, including MC-PixBiS~\cite{george2019deep}, MCCNN-OCCL-GMM~\cite{george2020learning}, MC-ResNetDLAS~\cite{parkin2019recognizing} and CMFL~\cite{george2021cross}. Compared with the previous best results, our MA-ViT achieves significantly better performance with a large margin in ``seen'' and ``unseen'' protocols: $-0.25\%$ over CMFL ($1.70\%$), and $-3.79\%$ over CMFL ($7.60\%$), respectively. It is worth noting that MA-ViT noticeably surpass these methods on two challenging sub-protocols of ``unseen'' protocol: when ``Flexiblemask'' and ``Glasses'' are not seen in the training stage. We analyze that the commonality of the two attacks is 3D facial structure, realistic color-texture, and only local regions contain spoofing traces, which are easy to be ignored by the ConvNet-based methods and play a long-term indicative role in ViT.

{ \textbf{Cross-Testing Results.}}
To evaluate the robustness, we conduct cross-testing experiments between models trained on MmFA and WMCA with Protocol ``seen''. 
We also introduce the previous SOTA ConvNet-based method Aux.(Depth)~\cite{liu2018learning} as a baseline. Tab.~\ref{tab:Cross_testing} lists the HTER of all methods trained on one dataset and tested on another dataset. From these results, MA-ViT outperforms Aux.(Depth) whether tested on WMCA ($24.54\%$ vs. $20.63\%$) or MmFA ($12.35\%$ vs. $10.41\%$). When comparing the results of MDA-ViT and CMA-ViT, we can conclude that MDA plays a more important role in improving the robustness. See Tab.~\ref{tab:Cross_testing} for details, when replacing the element of ViT from CMA to MDA, the results are further reduced from $27.97\%$ to $24.84\%$ when tested on WMCA and from $16.97\%$ to $13.30\%$ when tested on MmFA, respectively. Due to the mismatch of sensors, resolutions, attack types, and settings between different datasets, it is more effective to mine potential features by removing the interferences of modal information.
 \begin{table}[]
	\centering
	\resizebox{1.0\linewidth}{!}{
\begin{tabular}{|c|c|c|}
	\hline
	Method      & Train(MmFA),Test(WMCA) & Train(WMCA),Test(MmFA) \\ \hline
	Aux.(Depth) & 24.54                  & 12.35                  \\ \hline
	ViT         & 31.87                  & 22.79                  \\ \hline
	CMA-ViT     & 27.97                  & 16.97                  \\ \hline
	MDA-ViT     & 24.84                  & 13.30                  \\ \hline
	MA-ViT      & \textbf{20.63}         & \textbf{10.41}         \\ \hline
\end{tabular}
	}
	\caption{The HTER (\%) values from the cross-testing between MmFA and WMCA (Protocol ``seen") datasets.}
	\label{tab:Cross_testing}
\end{table}

\subsection{Flexible Modal Scenario Evaluations}  
In order to explore the ability of our approach for modality-agnostic features learning, we conducted experiments on dataset MmFA, and Protocol ``seen'' of WMCA. Before reporting the results in flexible modal scenarios, we first list the results of the SOTA methods on RGB (R), Depth (D) and IR (I) modalities on each dataset, \ie, MS-SEF for MmFA, and MC-CNN~\cite{george2019biometric} for WMCA.

By comparing the results in Tab.~\ref{tab:flexible_results}, we can draw the following two conclusions: (1) Our MA-ViT achieves the best results in each modality by training only one model. Such as compared with MS-SEF on MmFA, MA-ViT reduces the ACER of RGB, Depth and IR results by $9.57\%$ ($11.43\%$ vs. $21.00\%$), $2.80\%$ ($0.80\%$ vs. $3.60\%$) and $17.90\%$ ($1.50\%$ vs. $19.40\%$) points where they are trained in an independent model. Experimental results show that our model can flexibly test samples with any modalities without retraining the model again. (2) Our MATB can improve the any single modal performance with the help of multi-modal data. 
This clearly demonstrates that the proposed MATB enables any modal sequences to mine modality-agnostic liveness features for flexible modal testing. There is a similar conclusion on WMCA when compared with MC-CNN.
\begin{table}[]
	\centering
	\scalebox{0.65}{
\begin{tabular}{|ccccccccc|}
	\hline
	\multicolumn{1}{|c|}{\multirow{2}{*}{Method}}                                              & \multicolumn{1}{c|}{\multirow{2}{*}{Train}} & \multicolumn{1}{c|}{\multirow{2}{*}{Test}} & \multicolumn{3}{c|}{MmFA}                                                            & \multicolumn{3}{c|}{WMCA}                                       \\ \cline{4-9} 
	\multicolumn{1}{|c|}{}                                                                     & \multicolumn{1}{c|}{}                       & \multicolumn{1}{c|}{}                      & \multicolumn{1}{c|}{APCER} & \multicolumn{1}{c|}{BPCER} & \multicolumn{1}{c|}{ACER}  & \multicolumn{1}{c|}{APCER} & \multicolumn{1}{c|}{BPCER} & ACER  \\ \hline
	\multicolumn{9}{|c|}{Fixed modal testing}                                                                                                                                                                                                                                                                                                      \\ \hline
	\multicolumn{1}{|c|}{\multirow{3}{*}{\begin{tabular}[c]{@{}c@{}} SOTA \end{tabular}}} & \multicolumn{1}{c|}{R}                      & \multicolumn{1}{c|}{R}                     & \multicolumn{1}{c|}{40.30} & \multicolumn{1}{c|}{1.60}  & \multicolumn{1}{c|}{21.00} & \multicolumn{1}{c|}{65.65} & \multicolumn{1}{c|}{0.00}  & 32.82 \\ \cline{2-9} 
	\multicolumn{1}{|c|}{}                                                                     & \multicolumn{1}{c|}{D}                      & \multicolumn{1}{c|}{D}                     & \multicolumn{1}{c|}{6.00}  & \multicolumn{1}{c|}{1.20}  & \multicolumn{1}{c|}{3.60}  & \multicolumn{1}{c|}{11.77} & \multicolumn{1}{c|}{0.31}  & 6.04  \\ \cline{2-9} 
	\multicolumn{1}{|c|}{}                                                                     & \multicolumn{1}{c|}{I}                      & \multicolumn{1}{c|}{I}                     & \multicolumn{1}{c|}{38.60} & \multicolumn{1}{c|}{0.40}  & \multicolumn{1}{c|}{19.40} & \multicolumn{1}{c|}{5.03}  & \multicolumn{1}{c|}{0.00}  & 2.51  \\ \hline
	\multicolumn{1}{|c|}{\multirow{3}{*}{ViT}}                                                 & \multicolumn{1}{c|}{R}                      & \multicolumn{1}{c|}{R}       & \multicolumn{1}{c|}{22.67} & \multicolumn{1}{c|}{30.67} & \multicolumn{1}{c|}{26.67} & \multicolumn{1}{c|}{2.49}  & \multicolumn{1}{c|}{12.17} & 7.33  \\ \cline{2-9} 
	\multicolumn{1}{|c|}{}                                                                     & \multicolumn{1}{c|}{D}                      & \multicolumn{1}{c|}{D}                     & \multicolumn{1}{c|}{3.02}  & \multicolumn{1}{c|}{7.50}  & \multicolumn{1}{c|}{5.26}  & \multicolumn{1}{c|}{4.75}  & \multicolumn{1}{c|}{2.61}  & 3.68  \\ \cline{2-9} 
	\multicolumn{1}{|c|}{}                                                                     & \multicolumn{1}{c|}{I}                      & \multicolumn{1}{c|}{I}                     & \multicolumn{1}{c|}{3.52}  & \multicolumn{1}{c|}{9.50}  & \multicolumn{1}{c|}{6.51}  & \multicolumn{1}{c|}{4.98}  & \multicolumn{1}{c|}{4.35}  & 4.66  \\ \hline
	\multicolumn{9}{|c|}{Flexible modal testing}                                                                                                                                                                                                                                                                                                   \\ \hline
	\multicolumn{1}{|c|}{\multirow{3}{*}{MA-ViT}}                                              & \multicolumn{1}{c|}{R\&D\&I}                & \multicolumn{1}{c|}{R}                     & \multicolumn{1}{c|}{8.88}  & \multicolumn{1}{c|}{14.00} & \multicolumn{1}{c|}{11.43} & \multicolumn{1}{c|}{7.69}  & \multicolumn{1}{c|}{3.48}  & 5.59  \\ \cline{2-9} 
	\multicolumn{1}{|c|}{}                                                                     & \multicolumn{1}{c|}{R\&D\&I}                & \multicolumn{1}{c|}{D}                     & \multicolumn{1}{c|}{0.78}  & \multicolumn{1}{c|}{0.83}  & \multicolumn{1}{c|}{0.80}  & \multicolumn{1}{c|}{6.56}  & \multicolumn{1}{c|}{0.00}  & 3.28  \\ \cline{2-9} 
	\multicolumn{1}{|c|}{}                                                                     & \multicolumn{1}{c|}{R\&D\&I}                & \multicolumn{1}{c|}{I}                     & \multicolumn{1}{c|}{0.67}  & \multicolumn{1}{c|}{2.33}  & \multicolumn{1}{c|}{1.50}  & \multicolumn{1}{c|}{3.39}  & \multicolumn{1}{c|}{0.87}  & 2.13  \\ \hline
\end{tabular}
	}
	\caption{Comparison of flexible modal results (\%) based on multi-modal datasets. The `SOTA' means the method with public results on the corresponding dataset. R\&D\&I indicates the method receives RGB (R), Depth (D) and IR (I) paired samples as input.}
	\label{tab:flexible_results}
\end{table}

\subsection{Ablation Study}
{ \textbf{Effect of the Frameworks.}}
In this experiment, we compare MA-ViT(S) with a CNN method Aux.(Depth)~\cite{liu2018learning}, with a variety of ViT family architectures, \ie, baseline ViT(S), ``tiny'' (T) and ``base'' (B) variants. In addition to the metric of ACER, FLOPs and parameters measured for all the models and shown as Tab.~\ref{tab:ablation_frameworks}. Compared with ViT(S), our approach only introduces $0.35$ (G) FLOPs and $1.06$ (M) parameters, but has significant performance improvement on three modalities. When compared with MA-ViT(T) and MA-ViT(B), we find that the ``small'' variant is more suitable for face anti-spoofing tasks, which may be related to the dataset size and task category. Finally, it can be seen that MDA-ViT and CMA-ViT outperforms ViT(s) by a big margin, which improve the any single modal performance by eliminating modality-related information and fusing liveness features from another modal sequences. 
\begin{table}[]
	\centering
	\resizebox{1.0\linewidth}{!}{
		\begin{tabular}{|c|ccc|ccc|c|c|}
			\hline
			\multirow{2}{*}{Model} & \multicolumn{3}{c|}{MmFA}                                                               & \multicolumn{3}{c|}{WMCA}                                                               & \multirow{2}{*}{\begin{tabular}[c]{@{}c@{}}FLOPs\\ (G)\end{tabular}} & \multirow{2}{*}{\begin{tabular}[c]{@{}c@{}}Params\\ (M)\end{tabular}} \\ \cline{2-7}
			& \multicolumn{1}{c|}{R}             & \multicolumn{1}{c|}{D}             & I             & \multicolumn{1}{c|}{R}             & \multicolumn{1}{c|}{D}             & I             &                                                                      &                                                                       \\ \hline
			ViT(S)                    & \multicolumn{1}{c|}{26.67}         & \multicolumn{1}{c|}{5.26}          & 6.51          & \multicolumn{1}{c|}{7.33}          & \multicolumn{1}{c|}{3.68}          & 4.66          & 1.61                                                                 & 2.30                                                                  \\ \hline
			MDA-ViT($\lambda$=0.7)                & \multicolumn{1}{c|}{14.08}         & \multicolumn{1}{c|}{3.17}          & 1.75          & \multicolumn{1}{c|}{6.95}          & \multicolumn{1}{c|}{5.49}          & 6.60          & 1.92                                                                 & 3.36                                                                  \\ \hline
			MDA-ViT($\lambda$=0.8)                & \multicolumn{1}{c|}{13.78}         & \multicolumn{1}{c|}{2.01}          & 1.75          & \multicolumn{1}{c|}{6.25}          & \multicolumn{1}{c|}{2.61}          & 4.43          & 1.92                                                                 & 3.36                                                                  \\ \hline
			MDA-ViT($\lambda$=0.9)                & \multicolumn{1}{c|}{16.74}         & \multicolumn{1}{c|}{2.06}          & 2.22          & \multicolumn{1}{c|}{6.60}          & \multicolumn{1}{c|}{5.49}          & 4.76          & 1.92                                                                 & 3.36                                                                  \\ \hline
			CMA-ViT                & \multicolumn{1}{c|}{\textbf{9.71}} & \multicolumn{1}{c|}{3.25}          & 1.53          & \multicolumn{1}{c|}{8.97}          & \multicolumn{1}{c|}{\textbf{2.02}} & 2.13          & 1.64                                                                 & 2.30                                                                  \\ \hline
			MA-ViT(S)              & \multicolumn{1}{c|}{11.43}         & \multicolumn{1}{c|}{\textbf{0.80}} & \textbf{1.50} & \multicolumn{1}{c|}{\textbf{5.59}} & \multicolumn{1}{c|}{3.28}          & \textbf{2.13} & 1.96                                                                 & 3.36                                                                  \\ \hline
			MA-ViT(T)              & \multicolumn{1}{c|}{12.89}         & \multicolumn{1}{c|}{1.67}          & 3.64          & \multicolumn{1}{c|}{5.77}          & \multicolumn{1}{c|}{4.38}          & 2.92          & 0.49                                                                 & 0.86                                                                  \\ \hline
			MA-ViT(B)              & \multicolumn{1}{c|}{18.00}         & \multicolumn{1}{c|}{1.61}          & 5.31          & \multicolumn{1}{c|}{5.64}          & \multicolumn{1}{c|}{3.94}          & 2.70          & 7.79                                                                 & 13.26                                                                 \\ \hline
		\end{tabular}
	}
	\caption{Ablation study with different architecture on MmFA and WMCA with Protocol ``seen''. ACER values are reported.}
	\label{tab:ablation_frameworks}
\end{table}

{ \textbf{Effect of the $\lambda$ in MDA.}}
In MDA, the $\lambda$ determines the number of modality-irrelated patch tokens. The experimental results are shown in Tab.~\ref{tab:ablation_frameworks}. When $\lambda=0.8$ in MDA-ViT, all experiments achieve the best results.

\begin{figure}[t]
	\centering
	\includegraphics[width=1.0\linewidth]{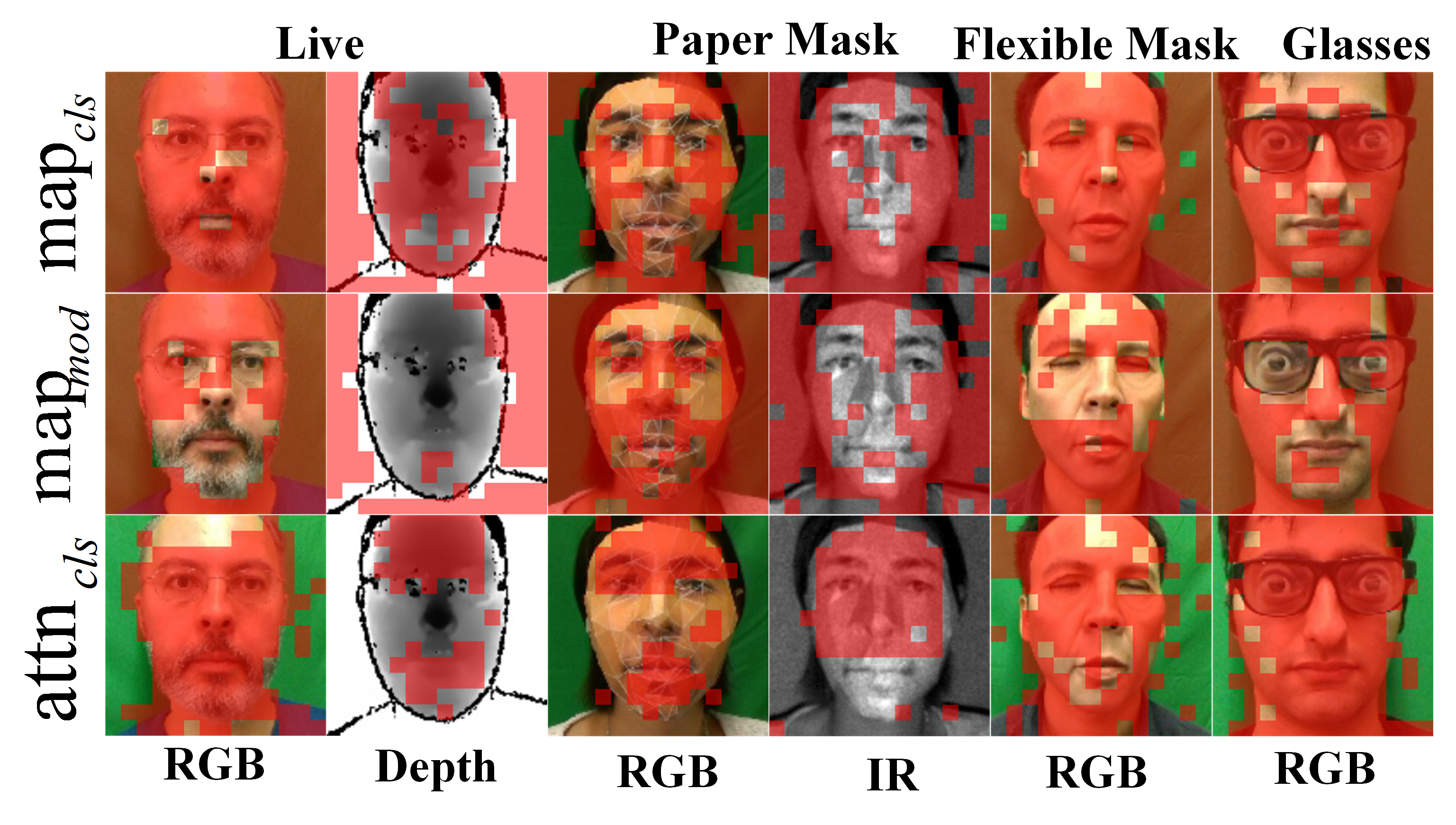}
	\caption{The mining process of informative patch tokens in each modal sequence which are covered by the red mask and obtained by thresholding the maps to keep 80\% ($\lambda=0.8$) of the mass. The sample is from the WMCA dataset, where the sample category is marked at the top.}
	\label{fig:mda_vis}
\end{figure}

In Fig.~\ref{fig:mda_vis}, we visualize the patch tokens mined by the MDA in the last layer. For each modal sequence (each column), we visualize the areas of modality-irrelated patch tokens (the first row), classification-irrelated patch tokens (the second row), and the final informative patch tokens (the third row). We can see that without MDA, the informative patch tokens are distributed in global regions, including modality-irrelated region and background region. While, the distribution is adjusted after MDA, which refers to the patch tokens in facial region.

\section{Conclusion}
In this work, we present a pure transformer-based framework named MA-ViT, which aims to guide any modal sequence to learn modality-agnostic liveness features, and fuse the liveness complementary information from other modalities through the developed MMDA and MCMA respectively.
Experiments show that our approach MA-ViT only introduces $0.35$ (G) FLOPs and $1.06$ (M) parameters, but gains significant improvements compared with baseline method ViT(S).


\section*{Acknowledgements}
This work was supported by the National Key R\&D Program of China 2021YFF0602103, the Science and Technology Development Fund of Macau (0008/2019/A1, 0010/2019/AFJ, 0025/2019/AKP, 0004/2020/A1, 0070/2021/AMJ) and Guangdong Provincial Key R\&D Programme: 2019B010148001.

\bibliographystyle{named}
\bibliography{ijcai22}

\begin{thebibliography}{}

\bibitem[\protect\citeauthoryear{Boulkenafet \bgroup \em et al.\egroup
  }{2017}]{Boulkenafet2017OULU}
Zinelabinde Boulkenafet, Jukka Komulainen, Lei Li, Xiaoyi Feng, and Abdenour
  Hadid.
\newblock Oulu-npu: A mobile face presentation attack database with real-world
  variations.
\newblock In {\em FGR}, pages 612--618, 2017.

\bibitem[\protect\citeauthoryear{Dosovitskiy \bgroup \em et al.\egroup
  }{2020}]{dosovitskiy2020image}
Alexey Dosovitskiy, Lucas Beyer, Alexander Kolesnikov, Dirk Weissenborn,
  Xiaohua Zhai, Thomas Unterthiner, Mostafa Dehghani, Matthias Minderer, Georg
  Heigold, Sylvain Gelly, et~al.
\newblock An image is worth 16x16 words: Transformers for image recognition at
  scale.
\newblock {\em arXiv preprint arXiv:2010.11929}, 2020.

\bibitem[\protect\citeauthoryear{George and Marcel}{2019}]{george2019deep}
Anjith George and S{\'e}bastien Marcel.
\newblock Deep pixel-wise binary supervision for face presentation attack
  detection.
\newblock In {\em ICB}, pages 1--8, 2019.

\bibitem[\protect\citeauthoryear{George and Marcel}{2020}]{george2020learning}
Anjith George and S{\'e}bastien Marcel.
\newblock Learning one class representations for face presentation attack
  detection using multi-channel convolutional neural networks.
\newblock {\em IEEE TIFS}, 16:361--375, 2020.

\bibitem[\protect\citeauthoryear{George and Marcel}{2021}]{george2021cross}
Anjith George and S{\'e}bastien Marcel.
\newblock Cross modal focal loss for rgbd face anti-spoofing.
\newblock In {\em CVPR}, pages 7882--7891, 2021.

\bibitem[\protect\citeauthoryear{George \bgroup \em et al.\egroup
  }{2019}]{george2019biometric}
Anjith George, Zohreh Mostaani, David Geissenbuhler, Olegs Nikisins, Andr{\'e}
  Anjos, and S{\'e}bastien Marcel.
\newblock Biometric face presentation attack detection with multi-channel
  convolutional neural network.
\newblock {\em IEEE TIFS}, 15:42--55, 2019.

\bibitem[\protect\citeauthoryear{Liu \bgroup \em et al.\egroup
  }{2016}]{liu20163d}
Siqi Liu, Baoyao Yang, Pong~C Yuen, and Guoying Zhao.
\newblock A 3d mask face anti-spoofing database with real world variations.
\newblock In {\em CVPRW}, pages 100--106, 2016.

\bibitem[\protect\citeauthoryear{Liu \bgroup \em et al.\egroup
  }{2018}]{liu2018learning}
Yaojie Liu, Amin Jourabloo, and Xiaoming Liu.
\newblock Learning deep models for face anti-spoofing: Binary or auxiliary
  supervision.
\newblock In {\em CVPR}, pages 389--398, 2018.

\bibitem[\protect\citeauthoryear{Liu \bgroup \em et al.\egroup
  }{2019}]{liu2019multi}
Ajian Liu, Jun Wan, Sergio Escalera, Hugo Jair~Escalante, Zichang Tan, Qi~Yuan,
  Kai Wang, Chi Lin, Guodong Guo, Isabelle Guyon, et~al.
\newblock Multi-modal face anti-spoofing attack detection challenge at
  cvpr2019.
\newblock In {\em CVPRW}, pages 0--0, 2019.

\bibitem[\protect\citeauthoryear{Liu \bgroup \em et al.\egroup
  }{2020}]{liu2020disentangling}
Yaojie Liu, Joel Stehouwer, and Xiaoming Liu.
\newblock On disentangling spoof trace for generic face anti-spoofing.
\newblock In {\em ECCV}, pages 406--422, 2020.

\bibitem[\protect\citeauthoryear{Liu \bgroup \em et al.\egroup
  }{2021a}]{liu2021cross}
Ajian Liu, Xuan Li, Jun Wan, Yanyan Liang, Sergio Escalera, Hugo~Jair
  Escalante, Meysam Madadi, Yi~Jin, Zhuoyuan Wu, Xiaogang Yu, et~al.
\newblock Cross-ethnicity face anti-spoofing recognition challenge: A review.
\newblock {\em IET Biometrics}, 10(1):24--43, 2021.

\bibitem[\protect\citeauthoryear{Liu \bgroup \em et al.\egroup
  }{2021b}]{liu2021casia}
Ajian Liu, Zichang Tan, Jun Wan, Sergio Escalera, Guodong Guo, and Stan~Z Li.
\newblock Casia-surf cefa: A benchmark for multi-modal cross-ethnicity face
  anti-spoofing.
\newblock In {\em WACV}, pages 1179--1187, 2021.

\bibitem[\protect\citeauthoryear{Liu \bgroup \em et al.\egroup
  }{2021c}]{liu2021face}
Ajian Liu, Zichang Tan, Jun Wan, Yanyan Liang, Zhen Lei, Guodong Guo, and
  Stan~Z Li.
\newblock Face anti-spoofing via adversarial cross-modality translation.
\newblock {\em IEEE TIFS}, 16:2759--2772, 2021.

\bibitem[\protect\citeauthoryear{Liu \bgroup \em et al.\egroup
  }{2021d}]{liu20213d}
Ajian Liu, Chenxu Zhao, Zitong Yu, Anyang Su, Xing Liu, Zijian Kong, Jun Wan,
  Sergio Escalera, Hugo~Jair Escalante, Zhen Lei, et~al.
\newblock 3d high-fidelity mask face presentation attack detection challenge.
\newblock In {\em ICCVW}, pages 814--823, 2021.

\bibitem[\protect\citeauthoryear{Liu \bgroup \em et al.\egroup
  }{2021e}]{liu2021contrastive}
Ajian Liu, Chenxu Zhao, Zitong Yu, Jun Wan, Anyang Su, Xing Liu, Zichang Tan,
  Sergio Escalera, Junliang Xing, Yanyan Liang, et~al.
\newblock Contrastive context-aware learning for 3d high-fidelity mask face
  presentation attack detection.
\newblock {\em arXiv preprint arXiv:2104.06148}, 2021.

\bibitem[\protect\citeauthoryear{Liu \bgroup \em et al.\egroup
  }{2021f}]{liu2021swin}
Ze~Liu, Yutong Lin, Yue Cao, Han Hu, Yixuan Wei, Zheng Zhang, Stephen Lin, and
  Baining Guo.
\newblock Swin transformer: Hierarchical vision transformer using shifted
  windows.
\newblock In {\em CVPR}, pages 10012--10022, 2021.

\bibitem[\protect\citeauthoryear{Parkin and
  Grinchuk}{2019}]{parkin2019recognizing}
Aleksandr Parkin and Oleg Grinchuk.
\newblock Recognizing multi-modal face spoofing with face recognition networks.
\newblock In {\em CVPRW}, pages 0--0, 2019.

\bibitem[\protect\citeauthoryear{Shao \bgroup \em et al.\egroup
  }{2019}]{shao2019multi}
Rui Shao, Xiangyuan Lan, Jiawei Li, and Pong~C Yuen.
\newblock Multi-adversarial discriminative deep domain generalization for face
  presentation attack detection.
\newblock In {\em CVPR}, pages 10023--10031, 2019.

\bibitem[\protect\citeauthoryear{Touvron \bgroup \em et al.\egroup
  }{2021}]{touvron2021training}
Hugo Touvron, Matthieu Cord, Matthijs Douze, Francisco Massa, Alexandre
  Sablayrolles, and Herv{\'e} J{\'e}gou.
\newblock Training data-efficient image transformers \& distillation through
  attention.
\newblock In {\em ICML}, pages 10347--10357, 2021.

\bibitem[\protect\citeauthoryear{Wang \bgroup \em et al.\egroup
  }{2021}]{wang2021kvt}
Pichao Wang, Xue Wang, Fan Wang, Ming Lin, Shuning Chang, Wen Xie, Hao Li, and
  Rong Jin.
\newblock Kvt: k-nn attention for boosting vision transformers.
\newblock {\em arXiv preprint arXiv:2106.00515}, 2021.

\bibitem[\protect\citeauthoryear{Xu \bgroup \em et al.\egroup
  }{}]{xu2021cdtrans}
Tongkun Xu, Weihua Chen, Pichao Wang, Fan Wang, Hao Li, and Rong Jin.
\newblock Cdtrans: Cross-domain transformer for unsupervised domain adaptation.
\newblock {\em arXiv preprint arXiv:2109.06165}.

\bibitem[\protect\citeauthoryear{Yu \bgroup \em et al.\egroup
  }{2020}]{yu2020searching}
Zitong Yu, Chenxu Zhao, Zezheng Wang, Yunxiao Qin, Zhuo Su, Xiaobai Li, Feng
  Zhou, and Guoying Zhao.
\newblock Searching central difference convolutional networks for face
  anti-spoofing.
\newblock In {\em CVPR}, pages 5295--5305, 2020.

\bibitem[\protect\citeauthoryear{Zhang \bgroup \em et al.\egroup
  }{2019}]{zhang2019dataset}
Shifeng Zhang, Xiaobo Wang, Ajian Liu, Chenxu Zhao, Jun Wan, Sergio Escalera,
  Hailin Shi, Zezheng Wang, and Stan~Z Li.
\newblock A dataset and benchmark for large-scale multi-modal face
  anti-spoofing.
\newblock In {\em CVPR}, pages 919--928, 2019.

\bibitem[\protect\citeauthoryear{Zhang \bgroup \em et al.\egroup
  }{2020a}]{zhang2020face}
Ke-Yue Zhang, Taiping Yao, Jian Zhang, Ying Tai, Shouhong Ding, Jilin Li,
  Feiyue Huang, Haichuan Song, and Lizhuang Ma.
\newblock Face anti-spoofing via disentangled representation learning.
\newblock In {\em ECCV}, pages 641--657, 2020.

\bibitem[\protect\citeauthoryear{Zhang \bgroup \em et al.\egroup
  }{2020b}]{zhang2020casia}
Shifeng Zhang, Ajian Liu, Jun Wan, Yanyan Liang, Guodong Guo, Sergio Escalera,
  Hugo~Jair Escalante, and Stan~Z Li.
\newblock Casia-surf: A large-scale multi-modal benchmark for face
  anti-spoofing.
\newblock {\em IEEE TBIOM}, 2(2):182--193, 2020.

\end{thebibliography}

\end{document}